\documentclass[11pt]{article}
\usepackage{acl07}
\usepackage{times}
\usepackage{latexsym}
\usepackage{haldefs}

\newcommand{\pitem}{\vspace{-0.2cm}\item}

\title{Frustratingly Easy Domain Adaptation}

\author{Hal Daum\'e III\\
        School of Computing\\
        University of Utah\\
        Salt Lake City, Utah 84112\\
        {\tt me@hal3.name}}

\date{}

\begin{document}
\maketitle

\begin{abstract}
We describe an approach to domain adaptation that is appropriate
exactly in the case when one has enough ``target'' data to do slightly
better than just using only ``source'' data.  Our approach is
incredibly simple, easy to implement as a preprocessing step (10 lines
of Perl!) and outperforms state-of-the-art approaches on a range of
datasets.  Moreover, it is trivially extended to a multi-domain
adaptation problem, where one has data from a variety of different
domains.
\end{abstract}

\section{Introduction} \label{sec:intro}

The task of domain adaptation is to develop learning algorithms that
can be easily ported from one domain to another---say, from newswire
to biomedical documents.  This problem is particularly interesting in
NLP because we are often in the situation that we have a large
collection of labeled data in one ``source'' domain (say, newswire)
but truly desire a model that performs well in a second ``target''
domain.  The approach we present in this paper is based on the idea of
transforming the domain adaptation learning problem into a standard
supervised learning problem to which any standard algorithm may be
applied (eg., maxent, SVMs, etc.).  Our transformation is incredibly
simple: we augment the feature space of both the source and target
data and use the result as input to a standard learning algorithm.

There are roughly two varieties of the domain adaptation problem that
have been addressed in the literature: the fully supervised case and
the semi-supervised case.  The fully supervised case models the
following scenario.  We have access to a large, annotated corpus of
data from a source domain.  In addition, we spend a little money to
annotate a small corpus in the target domain.  We want to leverage
both annotated datasets to obtain a model that performs well on the
target domain.  The semi-supervised case is similar, but instead of
having a small annotated target corpus, we have a large but
\emph{unannotated} target corpus.  In this paper, we focus exclusively
on the fully supervised case.

One particularly nice property of our approach is that it is
incredibly easy to implement: the Appendix provides a $10$ line, $194$
character Perl script for performing the complete transformation
(available at \url{http://hal3.name/easyadapt.pl.gz}).  In addition to
this simplicity, our algorithm performs as well as (or, in some cases,
better than) current state of the art techniques.

\section{Problem Formalization and Prior Work} \label{sec:prior}

To facilitate discussion, we first introduce some notation.  Denote by
$\cX$ the input space (typically either a real vector or a binary
vector), and by $\cY$ the output space.  We will write $\cD^s$ to
denote the distribution over source examples and $\cD^t$ to denote the
distribution over target examples.  We assume access to a samples $D^s
\sim \cD^s$ of source examples from the source domain, and samples
$D^t \sim \cD^t$ of target examples from the target domain.  We will
assume that $D^s$ is a collection of $N$ examples and $D^t$ is a
collection of $M$ examples (where, typically, $N \gg M$).  Our goal is
to learn a function $h : \cX \fto \cY$ with low expected loss with
respect to the target domain.  For the purposes of discussion, we will
suppose that $\cX = \R^F$ and that $\cY = \{ -1, +1 \}$.  However,
most of the techniques described in this section (as well as our own
technique) are more general.

There are several ``obvious'' ways to attack the domain adaptation
problem without developing new algorithms.  Many of these are
presented and evaluated by \newcite{daume06megam}.  

\begin{description}
\pitem
The \textsc{SrcOnly} baseline ignores the target data and trains a
single model, only on the source data.

\pitem
The \textsc{TgtOnly} baseline trains a single model only on the
target data.

\pitem
The \textsc{All} baseline simply trains a standard learning algorithm
on the union of the two datasets.

\pitem
A potential problem with the \textsc{All} baseline is that if $N \gg
M$, then $D^s$ may ``wash out'' any affect $D^t$ might have.  We will
discuss this problem in more detail later, but one potential solution
is to re-weight examples from $D^s$.  For instance, if $N = 10 \times
M$, we may weight each example from the source domain by $0.1$.  The
next baseline, \textsc{Weighted}, is exactly this approach, with the
weight chosen by cross-validation.

\pitem
The \textsc{Pred} baseline is based on the idea of using the output of
the source classifier as a feature in the target classifier.
Specifically, we first train a \textsc{SrcOnly} model.  Then we run
the \textsc{SrcOnly} model on the target data (training,
development and test).  We use the predictions made by the
\textsc{SrcOnly} model as additional features and train a second
model on the target data, augmented with this new feature.

\pitem
In the \textsc{LinInt} baseline, we linearly interpolate the
predictions of the \textsc{SrcOnly} and the \textsc{TgtOnly}
models.  The interpolation parameter is adjusted based on target
development data.
\end{description}

These baselines are actually surprisingly difficult to beat.  To date,
there are two models that have successfully defeated them on a handful
of datasets.  The first model, which we shall refer to as the
\textsc{Prior} model, was first introduced by \newcite{chelba04adapt}.
The idea of this model is to use the \textsc{SrcOnly} model as a
\emph{prior} on the weights for a second model, trained on the target
data.  \newcite{chelba04adapt} describe this approach within the
context of a maximum entropy classifier, but the idea is more general.
In particular, for many learning algorithms (maxent, SVMs, averaged
perceptron, naive Bayes, etc.), one \emph{regularizes} the weight
vector toward zero.  In other words, all of these algorithms contain a
regularization term on the weights $\vec w$ of the form $\lambda
\norm{\vec w}_2^2$.  In the generalized \textsc{Prior} model, we
simply replace this regularization term with $\lambda \norm{\vec w -
\vec w^s}_2^2$, where $\vec w^s$ is the weight vector learned in the
\textsc{SrcOnly} model.\footnote{For the maximum entropy, SVM and
naive Bayes learning algorithms, modifying the regularization term is
simple because it appears explicitly.  For the perceptron algorithm,
one can obtain an equivalent regularization by performing standard
perceptron updates, but using $(\vec w + \vec w^s)\T \vec x$ for
making predictions rather than simply $\vec w \T \vec x$.}  In this
way, the model trained on the target data ``prefers'' to have weights
that are similar to the weights from the \textsc{SrcOnly} model,
unless the data demands otherwise.  \newcite{daume06megam} provide
empirical evidence on four datasets that the \textsc{Prior} model
outperforms the baseline approaches.

More recently, \newcite{daume06megam} presented an algorithm for
domain adaptation for maximum entropy classifiers.  The key idea of
their approach is to learn \emph{three} separate models.  One model
captures ``source specific'' information, one captures ``target
specific'' information and one captures ``general'' information.  The
distinction between these three sorts of information is made on a
\emph{per-example} basis.  In this way, each source example is
considered either source specific or general, while each target
example is considered either target specific or general.
\newcite{daume06megam} present an EM algorithm for training their
model.  This model consistently outperformed all the baseline
approaches as well as the \textsc{Prior} model.  Unfortunately,
despite the empirical success of this algorithm, it is quite complex
to implement and is roughly $10$ to $15$ times slower than training
the \textsc{Prior} model.

\section{Adaptation by Feature Augmentation} \label{sec:adapt}

In this section, we describe our approach to the domain adaptation
problem.  Essentially, all we are going to do is take each feature in
the original problem and make three versions of it: a general version,
a source-specific version and a target-specific version.  The
augmented source data will contain only general and source-specific
versions.  The augmented target data contains general and
target-specific versions.

To state this more formally, first recall the notation from
Section~\ref{sec:prior}: $\cX$ and $\cY$ are the input and output
spaces, respectively; $D^s$ is the source domain data set and $D^t$ is
the target domain data set.  Suppose for simplicity that $\cX = \R^F$
for some $F > 0$.  We will define our augmented input space by $\breve
\cX = \R^{3F}$.  Then, define mappings $\Phi^s, \Phi^t : \cX \fto
\breve \cX$ for mapping the source and target data respectively.
These are defined by Eq~\eqref{eq:phis}, where $\vec 0 = \langle
0, 0, \dots, 0 \rangle \in \R^F$ is the zero vector.

\begin{equation}
\Phi^s(\vec x) = \langle \vec x , \vec x , \vec 0 \rangle
,\quad
\Phi^t(\vec x) = \langle \vec x , \vec 0 , \vec x \rangle
 \label{eq:phis}
\end{equation}

Before we proceed with a formal analysis of this transformation, let
us consider why it might be expected to work.  Suppose our task is
part of speech tagging, our source domain is the Wall Street Journal
and our target domain is a collection of reviews of computer hardware.
Here, a word like ``the'' should be tagged as a determiner in both
cases.  However, a word like ``monitor'' is more likely to be a verb
in the WSJ and more likely to be a noun in the hardware corpus.
Consider a simple case where $\cX = \R^2$, where $x_1$ indicates if
the word is ``the'' and $x_2$ indicates if the word is ``monitor.''
Then, in $\breve \cX$, $\breve x_1$ and $\breve x_2$ will be
``general'' versions of the two indicator functions, $\breve x_3$ and
$\breve x_4$ will be source-specific versions, and $\breve x_5$ and
$\breve x_6$ will be target-specific versions.

Now, consider what a learning algorithm could do to capture the fact
that the appropriate tag for ``the'' remains constant across the
domains, and the tag for ``monitor'' changes.  In this case, the model
can set the ``determiner'' weight vector to something like $\langle
1,0,0,0,0,0 \rangle$.  This places high weight on the common version
of ``the'' and indicates that ``the'' is most likely a determiner,
regardless of the domain.  On the other hand, the weight vector for
``noun'' might look something like $\langle 0,0,0,0,0,1 \rangle$,
indicating that the word ``monitor'' is a noun \emph{only} in the
target domain.  Similar, the weight vector for ``verb'' might look
like $\langle 0,0,0,1,0,0 \rangle$, indicating the ``monitor'' is a
verb \emph{only} in the source domain.

Note that this expansion is actually redundant.  We could equally well
use $\Phi^s(\vec x) = \langle \vec x, \vec x \rangle$ and $\Phi^t(\vec
x) = \langle \vec x, \vec 0 \rangle$.  However, it turns out that it
is easier to analyze the first case, so we will stick with that.
Moreover, the first case has the nice property that it is
straightforward to generalize it to the multi-domain adaptation
problem: when there are more than two domains.  In general, for $K$
domains, the augmented feature space will consist of $K+1$ copies of
the original feature space.

\subsection{A Kernelized Version}

It is straightforward to derive a kernelized version of the above
approach.  We do not exploit this property in our experiments---all
are conducted with a simple linear kernel.  However, by deriving the
kernelized version, we gain some insight into the method.  For this
reason, we sketch the derivation here.

Suppose that the data points $x$ are drawn from a reproducing kernel
Hilbert space $\cX$ with kernel $K : \cX \times \cX \fto \cR$, with
$K$ positive semi-definite.  Then, $K$ can be written as the dot
product (in $\cX$) of two (perhaps infinite-dimensional) vectors:
$K(x,x') = \langle \Phi(x), \Phi(x') \rangle_\cX$.  Define $\Phi^s$
and $\Phi^t$ in terms of $\Phi$, as:

\begin{align}
\Phi^s(x) &= \langle \Phi(x), \Phi(x), \vec 0 \rangle
 \label{eq:kphis} \\
\Phi^t(x) &= \langle \Phi(x), \vec 0 , \Phi(x)\rangle
 \nonumber
\end{align}

Now, we can compute the kernel product between $\Phi^s$ and $\Phi^t$
in the expanded RKHS by making use of the original kernel $K$.  We
denote the expanded kernel by $\breve K(x,x')$.  It is simplest to
first describe $\breve K(x,x')$ when $x$ and $x'$ are from the same
domain, then analyze the case when the domain differs.  When the
domain is the same, we get:
$\breve K(x,x')
= \langle \Phi(x), \Phi(x') \rangle_\cX + \langle \Phi(x), \Phi(x') \rangle_\cX 
= 2 K(x,x')$.  When they are from different domains, we get:
$\breve K(x,x')
= \langle \Phi(x), \Phi(x') \rangle_\cX
= K(x,x')$.  Putting this together, we have:

\begin{equation}
\breve K(x,x') =
\brack{ 2 K(x,x') & \text{same domain} \\
          K(x,x') & \text{diff. domain}}
\end{equation}

This is an intuitively pleasing result.  What it says is
that---considering the kernel as a measure of similarity---data points
from the same domain are ``by default'' twice as similar as those from
different domains.  Loosely speaking, this means that data points from
the target domain have twice as much influence as source points when
making predictions about test target data.

\subsection{Analysis}

We first note an obvious property of the feature-augmentation
approach.  Namely, it does not make learning harder, in a minimum
Bayes error sense.  A more interesting statement would be that it
makes learning \emph{easier}, along the lines of the result of
\cite{bendavid06adapt} --- note, however, that their results are for
the ``semi-supervised'' domain adaptation problem and so do not apply
directly.  As yet, we do not know a proper formalism in which to
analyze the fully supervised case.

It turns out that the feature-augmentation method is remarkably
similar to the \textsc{Prior} model\footnote{Thanks an anonymous
reviewer for pointing this out!}.  Suppose we learn feature-augmented
weights in a classifier regularized by an $\ell_2$ norm (eg., SVMs,
maximum entropy).  We can denote by $w_s$ the sum of the ``source''
and ``general'' components of the learned weight vector, and by $w_t$
the sum of the ``target'' and ``general'' components, so that $w_s$
and $w_t$ are the predictive weights for each task.  Then, the
regularization condition on the entire weight vector is approximately
$\norm{w_g}^2 + \norm{w_s-w_g}^2 + \norm{w_t-w_g}^2$, with free
parameter $w_g$ which can be chosen to minimize this sum.  This leads
to a regularizer proportional to $\norm{w_s-w_t}^2$, akin
to the \textsc{Prior} model.

Given this similarity between the feature-augmentation method and the
\textsc{Prior} model, one might wonder why we expect our approach to
do better.  Our belief is that this occurs because we optimize $w_s$
and $w_t$ \emph{jointly}, not sequentially.  First, this means that we
do not need to cross-validate to estimate good hyperparameters for
each task (though in our experiments, we do not use any
hyperparameters).  Second, and more importantly, this means that the
single supervised learning algorithm that is run is allowed to
regulate the trade-off between source/target and general weights.  In
the \textsc{Prior} model, we are forced to use the prior variance on
in the target learning scenario to do this ourselves.






\subsection{Multi-domain adaptation}

Our formulation is agnostic to the number of ``source'' domains.  In
particular, it may be the case that the source data actually falls
into a variety of more specific domains.  This is simple to account
for in our model.  In the two-domain case, we expanded the feature
space from $\R^F$ to $\R^{3F}$.  For a $K$-domain problem, we simply
expand the feature space to $\R^{(K+1)F}$ in the obvious way (the
``$+1$'' corresponds to the ``general domain'' while each of the other
$1\dots K$ correspond to a single task).

\section{Results}

In this section we describe experimental results on a wide variety of
domains.  First we describe the tasks, then we present experimental
results, and finally we look more closely at a few of the experiments.

\subsection{Tasks}

All tasks we consider are sequence labeling tasks (either named-entity
recognition, shallow parsing or part-of-speech tagging) on the
following datasets:

\begin{description}
\pitem{ACE-NER.}  We use data from the 2005 Automatic Content Extraction
  task, restricting ourselves to the named-entity recognition task.  The
  2005 ACE data comes from $5$ domains: Broadcast News (bn), Broadcast
  Conversations (bc), Newswire (nw), Weblog (wl), Usenet (un) and
  Converstaional Telephone Speech (cts).

\pitem{CoNLL-NE.}  Similar to ACE-NER, a named-entity recognition task.
  The difference is: we use the 2006 ACE data as the source domain
  and the CoNLL 2003 NER data as the target domain.

\pitem{PubMed-POS.}  A part-of-speech tagging problem on
  PubMed abstracts introduced by \newcite{blitzer06adaptation}.  There
  are two domains: the source domain is the WSJ portion of the
  Penn Treebank and the target domain is PubMed.

\pitem{CNN-Recap.}  This is a recapitalization task introduced by
  \newcite{chelba04adapt} and also used by \newcite{daume06megam}.
  The source domain is newswire and the target domain is the output of
  an ASR system.

\pitem{Treebank-Chunk.}  This is a shallow parsing task based on the
  data from the Penn Treebank.  This data comes from a variety of
  domains: the standard WSJ domain (we use the same data as for CoNLL
  2000), the ATIS switchboard domain, and the Brown corpus (which is,
  itself, assembled from six subdomains).

\pitem{Treebank-Brown.}  This is identical to the Treebank-Chunk task,
  except that we consider all of the Brown corpus to be a single
  domain.
\end{description}

In all cases (except for CNN-Recap), we use roughly the same feature
set, which has become somewhat standardized: lexical information
(words, stems, capitalization, prefixes and suffixes), membership on
gazetteers, etc.  For the CNN-Recap task, we use identical feature to
those used by both \newcite{chelba04adapt} and \newcite{daume06megam}:
the current, previous and next word, and 1-3 letter prefixes and
suffixes.

\begin{table}[t]
\small
\centering
\begin{tabular}{|l@{}c|rrr|r|}
\hline
{\bf Task}            & {\bf Dom   } & {\bf \# Tr} & {\bf \# De} & {\bf \# Te} & {\bf \# Ft} \\
\hline
                      & bn           & 52,998      & 6,625       & 6,626       & 80k         \\
                      & bc           & 38,073      & 4,759       & 4,761       & 109k        \\
ACE-                  & nw           & 44,364      & 5,546       & 5,547       & 113k        \\
NER                   & wl           & 35,883      & 4,485       & 4,487       & 109k        \\
                      & un           & 35,083      & 4,385       & 4,387       & 96k         \\
                      & cts          & 39,677      & 4,960       & 4,961       & 54k         \\
\hline
CoNLL-                & src          & 256,145     & -           & -           & 368k        \\
NER                   & tgt          & 29,791      & 5,258       & 8,806       & 88k         \\
\hline
PubMed-               & src          & 950,028     & -           & -           & 571k        \\
POS                   & tgt          &  11,264     & 1,987       & 14,554      & 39k         \\
\hline
CNN-                  & src          & 2,000,000   & -           & -           & 368k        \\
Recap                 & tgt          &    39,684   & 7,003       & 8,075       & 88k         \\
\hline
                      & wsj          & 191,209     & 29,455      & 38,440      & 94k         \\
                      & swbd3        & 45,282      & 5,596       & 41,840      & 55k         \\
                      & br-cf        & 58,201      & 8,307       & 7,607       & 144k        \\
Tree                  & br-cg        & 67,429      & 9,444       & 6,897       & 149k        \\
bank-                 & br-ck        & 51,379      & 6,061       & 9,451       & 121k        \\
Chunk                 & br-cl        & 47,382      & 5,101       & 5,880       & 95k         \\
                      & br-cm        & 11,696      & 1,324       & 1,594       & 51k         \\
                      & br-cn        & 56,057      & 6,751       & 7,847       & 115k        \\
                      & br-cp        & 55,318      & 7,477       & 5,977       & 112k        \\
                      & br-cr        & 16,742      & 2,522       & 2,712       & 65k         \\
\hline
\end{tabular}
\caption{Task statistics; columns are task, domain, size of the training,
development and test sets, and the number of unique features in the
training set.}
\label{tab:data}
\end{table}

Statistics on the tasks and datasets are in Table~\ref{tab:data}.

In all cases, we use the \textsc{Searn} algorithm for solving the
sequence labeling problem \cite{daume07searn} with an underlying
averaged perceptron classifier; implementation due to
\cite{daume04cg-bfgs}.  For structural features, we make a
second-order Markov assumption and only place a bias feature on the
transitions.  For simplicity, we optimize and report only on label
accuracy (but require that our outputs be parsimonious: we do not
allow ``I-NP'' to follow ``B-PP,'' for instance).  We do this for
three reasons.  First, our focus in this work is on building better
learning algorithms and introducing a more complicated measure only
serves to mask these effects.  Second, it is arguable that a measure
like $F_1$ is inappropriate for chunking tasks \cite{manning06f1}.
Third, we can easily compute statistical significance over accuracies
using McNemar's test.

\subsection{Experimental Results}

\begin{table*}[t]
\small
\centering
\begin{tabular}{|l@{}c|cccccccc|@{}c@{ }c@{}|}
\hline
{\bf Task}            & {\bf Dom} & \textsc{SrcOnly} & \textsc{TgtOnly} & \textsc{All} & \textsc{Weight} & \textsc{Pred} & \textsc{LinInt} & \textsc{Prior} & \textsc{Augment} & T$<$S & {\bf Win} \\
\hline
                      & bn           &     4.98  &     2.37  &     2.29  &     2.23  &     2.11  &     2.21  &     2.06  & \bf 1.98  & + & + \\
                      & bc           &     4.54  &     4.07  &     3.55  &     3.53  &     3.89  &     4.01  & \bf 3.47  & \bf 3.47  & + & + \\
ACE-                  & nw           &     4.78  &     3.71  &     3.86  &     3.65  &     3.56  &     3.79  &     3.68  & \bf 3.39  & + & + \\
NER                   & wl           &     2.45  &     2.45  & \bf 2.12  & \bf 2.12  &     2.45  &     2.33  &     2.41  & \bf 2.12  & = & + \\
                      & un           &     3.67  &     2.46  &     2.48  &     2.40  &     2.18  &     2.10  &     2.03  & \bf 1.91  & + & + \\
                      & cts          &     2.08  &     0.46  &     0.40  &     0.40  &     0.46  &     0.44  & \bf 0.34  & \bf 0.32  & + & + \\
\hline
CoNLL                 & tgt          &     2.49  &     2.95  &     1.80  & \bf 1.75  &     2.13  & \bf 1.77  &     1.89  & \bf 1.76  &   & + \\
\hline
PubMed                & tgt          &    12.02  &     4.15  &     5.43  &     4.15  &     4.14  &     3.95  &     3.99  & \bf 3.61  & + & + \\
\hline
CNN                   & tgt          &    10.29  &     3.82  &     3.67  &     3.45  &     3.46  &     3.44  & \bf 3.35  & \bf 3.37  & + & + \\
\hline
                      & wsj          &     6.63  &     4.35  &     4.33  &     4.30  &     4.32  &     4.32  &     4.27  & \bf 4.11  & + & + \\
                      & swbd3        &    15.90  &     4.15  &     4.50  &     4.10  &     4.13  &     4.09  &     3.60  & \bf 3.51  & + & + \\
                      & br-cf        &     5.16  &     6.27  &     4.85  &     4.80  &     4.78  & \bf 4.72  &     5.22  &     5.15  &   &   \\
Tree                  & br-cg        &     4.32  &     5.36  & \bf 4.16  & \bf 4.15  &     4.27  &     4.30  &     4.25  &     4.90  &   &   \\
bank-                 & br-ck        &     5.05  &     6.32  &     5.05  &     4.98  & \bf 5.01  & \bf 5.05  &     5.27  &     5.41  &   &   \\
Chunk                 & br-cl        &     5.66  &     6.60  &     5.42  & \bf 5.39  & \bf 5.39  &     5.53  &     5.99  &     5.73  &   &   \\
                      & br-cm        &     3.57  &     6.59  & \bf 3.14  & \bf 3.11  &     3.15  &     3.31  &     4.08  &     4.89  &   &   \\
                      & br-cn        &     4.60  &     5.56  &     4.27  &     4.22  & \bf 4.20  & \bf 4.19  &     4.48  &     4.42  &   &   \\
                      & br-cp        &     4.82  &     5.62  &     4.63  & \bf 4.57  & \bf 4.55  & \bf 4.55  &     4.87  &     4.78  &   &   \\
                      & br-cr        &     5.78  &     9.13  &     5.71  &     5.19  &     5.20  & \bf 5.15  &     6.71  &     6.30  &   &   \\
\hline
Treebank-             & brown        &     6.35  &     5.75  &     4.80  &     4.75  &     4.81  &     4.72  &     4.72  & \bf 4.65  & + & + \\
\hline
\end{tabular}
\caption{Task results.}
\label{tab:results}
\end{table*}

The full---somewhat daunting---table of results is presented in
Table~\ref{tab:results}.  The first two columns specify the task and
domain.  For the tasks with only a single source and target, we simply
report results on the target.  For the multi-domain adaptation tasks,
we report results for each setting of the target (where all other
data-sets are used as different ``source'' domains).  The next set of
eight columns are the \emph{error rates} for the task, using one of
the different techniques (``\textsc{Augment}'' is our proposed
technique).  For each row, the error rate of the best performing
technique is bolded (as are all techniques whose performance is not
statistically significantly different at the 95\% level).  The
``T$<$S'' column is contains a ``+'' whenever \textsc{TgtOnly}
outperforms \textsc{SrcOnly} (this will become important shortly).
The final column indicates when \textsc{Augment} comes
in first.\footnote{One advantage of using the averaged perceptron for
all experiments is that the only tunable hyperparameter is the number
of iterations.  In all cases, we run $20$ iterations and choose the
one with the lowest error on development data.}

There are several trends to note in the results.  Excluding for a
moment the ``br-*'' domains on the Treebank-Chunk task, our technique
always performs best.  Still excluding ``br-*'', the clear
second-place contestant is the \textsc{Prior} model, a finding
consistent with prior research.  When we repeat the Treebank-Chunk
task, but lumping all of the ``br-*'' data together into a single
``brown'' domain, the story reverts to what we expected before: our
algorithm performs best, followed by the \textsc{Prior} method.

Importantly, this simple story breaks down on the Treebank-Chunk task
for the eight sections of the Brown corpus.  For these, our
\textsc{Augment} technique performs rather poorly.  Moreover, there is
no clear winning approach on this task.  Our hypothesis is that the
common feature of these examples is that these are exactly the tasks
for which \textsc{SrcOnly} outperforms \textsc{TgtOnly} (with one
exception: CoNLL).  This seems like a plausible explanation, since it
implies that the source and target domains may not be that different.
If the domains are so similar that a large amount of source data
outperforms a small amount of target data, then it is unlikely that
blowing up the feature space will help.

We additionally ran the \textsc{MegaM} model \cite{daume06megam} on
these data (though not in the multi-conditional case; for this, we
considered the single source as the union of all sources).  The
results are not displayed in Table~\ref{tab:results} to save space.
For the majority of results, \textsc{MegaM} performed roughly
comparably to the best of the systems in the table.  In particular, it
was not statistically significantly different that \textsc{Augment}
on: ACE-NER, CoNLL, PubMed, Treebank-chunk-wsj, Treebank-chunk-swbd3,
CNN and Treebank-brown.  It did outperform \textsc{Augment} on the
Treebank-chunk on the Treebank-chunk-br-* data sets, but only
outperformed the best other model on these data sets for br-cg, br-cm
and br-cp.  However, despite its advantages on these data sets, it was
quite significantly slower to train: a single run required about ten
times longer than any of the other models (including
\textsc{Augment}), and also required five-to-ten iterations of
cross-validation to tune its hyperparameters so as to achieve these
results.



\subsection{Model Introspection}

One explanation of our model's improved performance is simply that by
augmenting the feature space, we are creating a more powerful model.
While this may be a partial explanation, here we show that what the
model learns about the various domains actually makes some plausible
sense.

We perform this analysis only on the ACE-NER data by looking
specifically at the learned weights.  That is, for any given feature
$f$, there will be seven versions of $f$: one corresponding to the
``cross-domain'' $f$ and seven corresponding to each domain.  We
visualize these weights, using Hinton diagrams, to see how the weights
vary across domains.

\begin{figure}[t]
\begin{tabular}{p{.7cm}p{6.5cm}}
&
\hspace{3.5mm}*\hspace{6mm}
bn\hspace{5mm}
bc\hspace{4.2mm}
nw\hspace{4.2mm}
wl\hspace{5mm}
un\hspace{4.2mm}
cts
\\
\vspace{-33.5mm}PER\vspace{4.5mm}
GPE\vspace{4.5mm}
ORG\vspace{4.5mm}
LOC
&
\psfig{figure=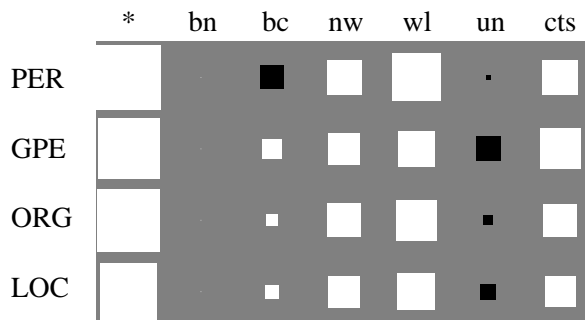,width=6.5cm}
\end{tabular}
\vspace{-8mm}
\caption{Hinton diagram for feature /Aa+/ at current position.}
\label{fig:hinton-cap}
\end{figure}

For example, consider the feature ``current word has an initial
capital letter and is then followed by one or more lower-case
letters.''  This feature is presumably useless for data that lacks
capitalization information, but potentially quite useful for other
domains.  In Figure~\ref{fig:hinton-cap} we shown a Hinton diagram for
this figure.  Each column in this figure correspond to a domain (the
top row is the ``general domain'').  Each row corresponds to a
class.\footnote{Technically there are many more classes than are shown
here.  We do not depict the smallest classes, and have merged the
``Begin-*'' and ``In-*'' weights for each entity type.}  Black boxes
correspond to negative weights and white boxes correspond to positive
weights.  The size of the box depicts the absolute value of the
weight.

As we can see from Figure~\ref{fig:hinton-cap}, the /Aa+/ feature is a
very good indicator of entity-hood (it's value is strongly positive
for all four entity classes), regardless of domain (i.e., for the
``*'' domain).  The lack of boxes in the ``bn'' column means that,
beyond the settings in ``*'', the broadcast news is agnostic with
respect to this feature.  This makes sense: there is no capitalization
in broadcast news domain, so there would be no sense is setting these
weights to anything by zero.  The usenet column is filled with
negative weights.  While this may seem strange, it is due to the fact
that many email addresses and URLs match this pattern, but are not
entities.

\begin{figure}[t]
\begin{tabular}{p{.7cm}p{6.5cm}}
&
\hspace{3.5mm}*\hspace{6mm}
bn\hspace{5mm}
bc\hspace{4.2mm}
nw\hspace{4.2mm}
wl\hspace{5mm}
un\hspace{4.2mm}
cts
\\
\vspace{-33.5mm}PER\vspace{4.5mm}
GPE\vspace{4.5mm}
ORG\vspace{4.5mm}
LOC
&
\psfig{figure=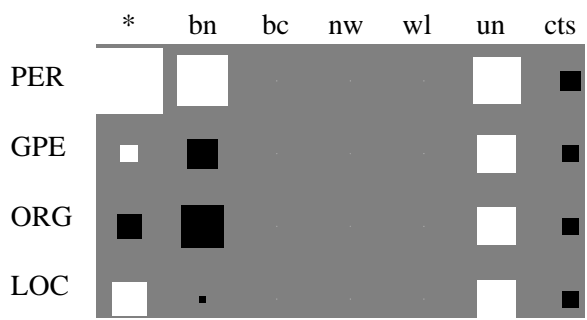,width=6.5cm}
\end{tabular}
\vspace{-8mm}
\caption{Hinton diagram for feature /bush/ at current position.}
\label{fig:hinton-bush}
\end{figure}

Figure~\ref{fig:hinton-bush} depicts a similar figure for the feature
``word is 'bush' at the current position'' (this figure is case
sensitive).\footnote{The scale of weights across features is
\emph{not} comparable, so do not try to compare
Figure~\ref{fig:hinton-cap} with Figure~\ref{fig:hinton-bush}.}  These
weights are somewhat harder to interpret.  What is happening is that
``by default'' the word ``bush'' is going to be a person---this is
because it rarely appears referring to a plant and so even in the
capitalized domains like broadcast conversations, if it appears at
all, it is a person.  The exception is that in the conversations data,
people \emph{do} actually talk about bushes as plants, and so the
weights are set accordingly.  The weights are high in the usenet
domain because people tend to talk about the president without
capitalizing his name.

\begin{figure}[t]
\begin{tabular}{p{.7cm}p{6.5cm}}
&
\hspace{3.5mm}*\hspace{6mm}
bn\hspace{5mm}
bc\hspace{4.2mm}
nw\hspace{4.2mm}
wl\hspace{5mm}
un\hspace{4.2mm}
cts
\\
\vspace{-33.5mm}PER\vspace{4.5mm}
GPE\vspace{4.5mm}
ORG\vspace{4.5mm}
LOC
&
\psfig{figure=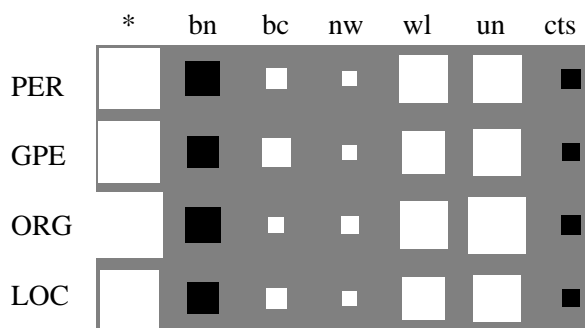,width=6.5cm}
\end{tabular}
\vspace{-8mm}
\caption{Hinton diagram for feature /the/ at current position.}
\label{fig:hinton-the}
\end{figure}

Figure~\ref{fig:hinton-the} presents the Hinton diagram for the
feature ``word at the current position is 'the''' (again,
case-sensitive).  In general, it appears, ``the'' is a common word in
entities in all domain except for broadcast news and conversations.
The exceptions are broadcast news and conversations.  These exceptions
crop up because of the capitalization issue.

\begin{figure}[t]
\begin{tabular}{p{.7cm}p{6.5cm}}
&
\hspace{3.5mm}*\hspace{6mm}
bn\hspace{5mm}
bc\hspace{4.2mm}
nw\hspace{4.2mm}
wl\hspace{5mm}
un\hspace{4.2mm}
cts
\\
\vspace{-33.5mm}PER\vspace{4.5mm}
GPE\vspace{4.5mm}
ORG\vspace{4.5mm}
LOC
&
\psfig{figure=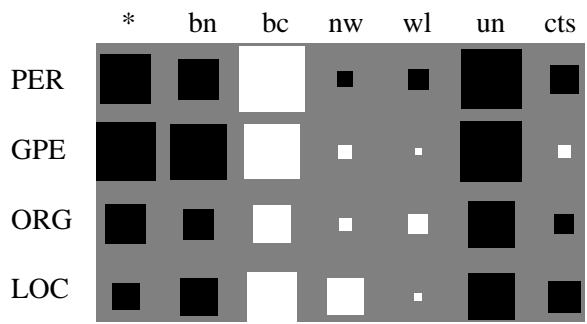,width=6.5cm}
\end{tabular}
\vspace{-8mm}
\caption{Hinton diagram for feature /the/ at previous position.}
\label{fig:hinton-the-1}
\end{figure}

In Figure~\ref{fig:hinton-the-1}, we show the diagram for the
feature ``previous word is 'the'.''  The only domain
for which this is a good feature of entity-hood is broadcast
conversations (to a much lesser extent, newswire).  This
occurs because of four phrases very common in the broadcast
conversations and rare elsewhere: ``the Iraqi people''
(``Iraqi'' is a GPE), ``the Pentagon'' (an ORG), ``the Bush
(cabinet$\vert$advisors$\vert$\dots)'' (PER), and ``the South'' (LOC).

\begin{figure}[t]
\begin{tabular}{p{.7cm}p{6.5cm}}
&
\hspace{3.5mm}*\hspace{6mm}
bn\hspace{5mm}
bc\hspace{4.2mm}
nw\hspace{4.2mm}
wl\hspace{5mm}
un\hspace{4.2mm}
cts
\\
\vspace{-33.5mm}PER\vspace{4.5mm}
GPE\vspace{4.5mm}
ORG\vspace{4.5mm}
LOC
&
\psfig{figure=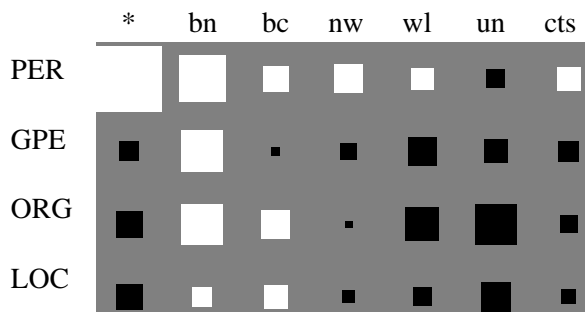,width=6.5cm}
\end{tabular}
\vspace{-8mm}
\caption{Hinton diagram for membership on a list of names at current position.}
\label{fig:hinton-names}
\end{figure}

Finally, Figure~\ref{fig:hinton-names} shows the Hinton diagram for
the feature ``the current word is on a list of common names'' (this
feature is case-\emph{in}sensitive).  All around, this is a good
feature for picking out people and nothing else.  The two exceptions
are: it is also a good feature for other entity types for broadcast
news and it is not quite so good for people in usenet.  The first is
easily explained: in broadcast news, it is very common to refer to
countries and organizations by the name of their respective leaders.
This is essentially a metonymy issue, but as the data is annotated,
these are marked by their true referent.  For usenet, it is because
the list of names comes from news data, but usenet names are more
diverse.

In general, the weights depicte for these features make some intuitive
sense (in as much as weights for any learned algorithm make intuitive
sense).  It is particularly interesting to note that while there are
some regularities to the patterns in the five diagrams, it is
definitely \emph{not} the case that there are, eg., two domains that
behave identically across all features.  This supports the hypothesis
that the reason our algorithm works so well on this data is because
the domains are actually quite well separated.


\section{Discussion}

In this paper we have described an \emph{incredibly} simple approach
to domain adaptation that---under a common and easy-to-verify
condition---outperforms previous approaches.  While it is somewhat
frustrating that something so simple does so well, it is perhaps not
surprising.  By augmenting the feature space, we are essentially
forcing the learning algorithm to do the adaptation for us.  Good
supervised learning algorithms have been developed over decades, and
so we are essentially just leveraging all that previous work.  Our
hope is that this approach is so simple that it can be used for many
more real-world tasks than we have presented here with little effort.
Finally, it is very interesting to note that using our method, shallow
parsing error rate on the CoNLL section of the treebank improves from
$5.35$ to $5.11$.  While this improvement is small, it is real, and
may carry over to full parsing.  The most important avenue of future
work is to develop a formal framework under which we can analyze this
(and other supervised domain adaptation models) theoretically.
Currently our results only state that this augmentation procedure
doesn't make the learning harder --- we would like to know that it
actually makes it easier.  An additional future direction is to
explore the kernelization interpretation further: why should we use
$2$ as the ``similarity'' between domains---we could introduce a
hyperparamter $\al$ that indicates the similarity between domains and
could be tuned via cross-validation.






\paragraph{Acknowledgments.}  We thank the three anonymous
reviewers, as well as Ryan McDonald and John Blitzer for very helpful
comments and insights.

\begin{small}
\bibliographystyle{acl}
\bibliography{bibfile}

\begin{thebibliography}{}

\bibitem[\protect\citename{Ben-David \bgroup et al.\egroup
  }2006]{bendavid06adapt}
Shai Ben-David, John Blitzer, Koby Crammer, and Fernando Pereira.
\newblock 2006.
\newblock Analysis of representations for domain adaptation.
\newblock In {\em Advances in Neural Information Processing Systems (NIPS)}.

\bibitem[\protect\citename{Blitzer \bgroup et al.\egroup
  }2006]{blitzer06adaptation}
John Blitzer, Ryan McDonald, and Fernando Pereira.
\newblock 2006.
\newblock Domain adaptation with structural correspondence learning.
\newblock In {\em Proceedings of the Conference on Empirical Methods in Natural
  Language Processing (EMNLP)}.

\bibitem[\protect\citename{Chelba and Acero}2004]{chelba04adapt}
Ciprian Chelba and Alex Acero.
\newblock 2004.
\newblock Adaptation of maximum entropy classifier: Little data can help a lot.
\newblock In {\em Proceedings of the Conference on Empirical Methods in Natural
  Language Processing (EMNLP)}, Barcelona, Spain.

\bibitem[\protect\citename{{Daum\'e III} and Marcu}2006]{daume06megam}
Hal {Daum\'e III} and Daniel Marcu.
\newblock 2006.
\newblock Domain adaptation for statistical classifiers.
\newblock {\em Journal of Artificial Intelligence Research}, 26.

\bibitem[\protect\citename{{Daum\'e III} \bgroup et al.\egroup
  }2007]{daume07searn}
Hal {Daum\'e III}, John Langford, and Daniel Marcu.
\newblock 2007.
\newblock Search-based structured prediction.
\newblock {\em Machine Learning Journal (submitted)}.

\bibitem[\protect\citename{{Daum\'e III}}2004]{daume04cg-bfgs}
Hal {Daum\'e III}.
\newblock 2004.
\newblock Notes on {CG} and {LM-BFGS} optimization of logistic regression.
\newblock Paper available at \url{http://pub.hal3.name/#daume04cg-bfgs},
  implementation available at \url{http://hal3.name/megam/}, August.

\bibitem[\protect\citename{Manning}2006]{manning06f1}
Christopher Manning.
\newblock 2006.
\newblock Doing named entity recognition? {Don't} optimize for {F$_1$}.
\newblock Post on the NLPers Blog, 25 August.
\newblock
  \url{http://nlpers.blogspot.com/2006/08/doing-named-entity-recognition-dont.%
html}.

\end{thebibliography}
\end{small}

\end{document}